%% file: main.tex
\documentclass[runningheads]{llncs}
\usepackage{graphicx}

\usepackage{tikz}
\usepackage{comment}
\usepackage{amsmath,amssymb} 
\usepackage{color}
\usepackage[pagebackref=true,breaklinks=true,colorlinks,bookmarks=false]{hyperref}
\usepackage{caption}
\usepackage{subcaption}
\usepackage{orcidlink}

\usepackage[accsupp]{axessibility}  

\usepackage{pifont}
\newcommand{\cmark}{\ding{51}}
\newcommand{\xmark}{\ding{55}}

\usepackage{booktabs}
\usepackage{multirow}
\begin{document}
\pagestyle{headings}
\mainmatter
\def\ECCVSubNumber{14}  

\title{Doc2Graph: a Task Agnostic Document Understanding Framework based on Graph Neural Networks} 

\titlerunning{Doc2Graph: a Document Understanding Framework based on GNNs}
%
\author{Andrea Gemelli\inst{1}\orcidlink{0000-0002-6149-8282} \and
Sanket Biswas\inst{2}\orcidlink{0000-0001-6648-8270} \and
Enrico Civitelli\inst{1}\orcidlink{0000-0001-5322-4831} \and
Josep Lladós\inst{2}\orcidlink{0000-0002-4533-4739} \and
Simone Marinai\inst{1}\orcidlink{0000-0002-6702-2277}}
\authorrunning{A. Gemelli et al.}
\institute{Dipartimento di Ingegneria dell'Informazione (DINFO)\\
Università degli studi di Firenze, Italy\\
\email{\{andrea.gemelli, enrico.civitelli, simone.marinai\}@unifi.it}
\and
Computer Vision Center \& Computer Science Department\\
Universitat Autònoma de Barcelona, Spain\\
\email{\{sbiswas, josep\}@cvc.uab.es}}

\maketitle

\begin{abstract}
Geometric Deep Learning has recently attracted significant
interest in a wide range of machine learning fields, including document analysis. The application of Graph Neural Networks (GNNs) has become crucial in various document-related tasks since they can unravel important structural patterns, fundamental in key information extraction processes. Previous works in the literature propose task-driven models and do not take into account the full power of graphs. We propose Doc2Graph, a task-agnostic document understanding framework based on a GNN model, to solve different tasks given different types of documents. We evaluated our approach on two challenging datasets for key information extraction in form understanding, invoice layout analysis and table detection. Our code is freely accessible on \url{https://github.com/andreagemelli/doc2graph}.

\end{abstract}

\section{Introduction}\label{s:intro}
\input{./tex/intro.tex}

\section{Related Work}\label{s:sota}
\input{./tex/sota.tex}

\section{Method}\label{s:method}
\input{./tex/method.tex}

\section{Experiments and Results}\label{s:results}
\input{./tex/results.tex}

\section{Conclusion}\label{s:conclusion}
\input{./tex/conclusion.tex}

\section*{Acknowledgment}

This work has been partially supported by the the Spanish projects MIRANDA RTI2018-095645-B-C21 and GRAIL PID2021-126808OB-I00, the CERCA Program / Generalitat de Catalunya, the FCT-19-15244, and PhD Scholarship from AGAUR (2021FIB-10010).

\clearpage
%
%
\bibliographystyle{splncs04}
\bibliography{main}

\end{document}

%% file: tex/intro.tex
Document Intelligence deals with the ability to read, understand and interpret documents.
Document understanding can be backed by graph representations, that robustly represent objects and relations. Graph reasoning for document parsing involves manipulating structured representations of semantically meaningful document objects (titles, tables, figures) and relations, using compositional rules. Customarily,  graphs have been selected as an adequate framework for leveraging structural information from documents, due to their inherent representational power to codify the object components (or semantic entities) and their pairwise relationships. In this context, recently graph neural networks (GNNs) have emerged as a powerful tool to tackle the problems of Key Information Extraction (KIE) ~\cite{carbonell2021named,yu2021pick}, Document Layout Analysis (DLA) which includes well-studied sub-tasks like table detection~\cite{riba2019table,riba2022table}, table structure recognition~\cite{liu2022neural,xue2021tgrnet} and table extraction~\cite{gemelli-icpr}, Visual Question Answering (VQA)~\cite{liang2021multi,li2022text}, synthetic document generation~\cite{biswas2021graph} and so on. 

Simultaneously, the common state-of-the-art practice in the document understanding community is to utilize the power of huge pre-trained vision-language models~\cite{appalaraju2021docformer,xu2020layoutlmv2,xu2020layoutlm} that learn whether the visual, textual and layout cues of the document are correlated. Despite achieving superior performance on most document understanding tasks, large-scale document pre-training comes with a high computational cost both in terms of memory and training time. 
We present a solution that does not rely on huge vision-language model pre-training modules, but rather recognizes the semantic text entities and their relationships from documents exploiting graphs. The solution has experimented on two challenging benchmarks for forms~\cite{FUNSD} and invoices~\cite{goldmann_lutz_2019_3257319} with a very small amount of labeled training data. 

Inspired by some prior works~\cite{davis2021visual,riba2019table,riba2022table}, we introduce \emph{Doc2Graph}, a novel task-agnostic framework to exploit graph-based representations for document understanding. The proposed model is validated in three different challenges, namely KIE in form understanding, invoice layout analysis and table detection. 
A graph representation module is proposed to organize the document objects. The graph nodes represent words or the semantic entities while edges the pairwise relationships between them. Finding the optimal set of edges to create the graph is anything but trivial: usually in literature heuristics are applied, e.g. using a visibility graph~\cite{riba2019table}. In this work, we do not make any assumption a priori on the connectivity: rather we attempt to build a fully connected graph representation over documents and let the network learn by itself what is relevant.

In summary, the primary contributions of this work can be summarized as follows:
\begin{itemize}
    \item Doc2Graph, the first task-agnostic GNN-based document understanding framework, evaluated on two challenging benchmarks (form and invoice understanding) for three significant tasks, without any requirement of huge pre-training data;
    \item We propose a general graph representation module for documents, that do not rely on heuristics to build pairwise relationships between words or entities;
    \item A novel GNN architectural pipeline with node and edge aggregation functions suited for documents, that exploits the relative positioning of document objects through polar coordinates.
\end{itemize}   

The rest of the paper is organized as follows. In section \ref{s:sota} we review the state-of-the-art in graph representation learning and vision-language models for document understanding. Section \ref{s:method} provides the details of the main methodological contribution. The experimental evaluation is reported in section \ref{s:results}. Finally, the conclusions are drawn in section \ref{s:conclusion}.

%% file: tex/sota.tex
Document understanding has been studied extensively in the last few years, owing to the advent of deep learning, but has been reformulated in a recent survey by Borchmann et. al.~\cite{borchmann2021due}. 
The tasks range from KIE performed for understanding forms~\cite{FUNSD}, receipts~\cite{huang2019icdar2019} and invoices~\cite{goldmann_lutz_2019_3257319}, to multimodal comprehension of both visual and textual cues in a document for classification~\cite{xu2020layoutlmv2,xu2020layoutlm}. It also includes the DLA task where recent works focus on building an end-to-end framework for both detection and classification of page regions~\cite{biswas2021beyond,biswas2022docsegtr}. Table detection~\cite{riba2019table,riba2022table}, structure recognition~\cite{liu2022neural,raja2020table} and extraction~\cite{gemelli-icpr,smock2022pubtables} in DLA gathered some special attention in recent years due to the high variability of layouts that make the both necessary to be solved and challenging to be tackled. In addition, question answering~\cite{mathew2021docvqa,singh2019towards} has emerged as an extension of the KIE task principle, where a natural language question replaces a property name. Current state-of-the-art approaches~\cite{appalaraju2021docformer,hong2020bros,powalski2021going,xu2020layoutlmv2,xu2020layoutlm} on these document understanding tasks have utilized the power of large pre-trained language models, relying on language more than the visual and geometrical information in a document and also end up using hundreds of millions of parameters in the process. Moreover, most of these models are trained with a huge transformer pipeline, which requires an immense amount of data during pre-training. In this regard, Davis et al.~\cite{davis2019deep} and Sarkar et al.~\cite{sarkar2020document} proposed language-agnostic models. In ~\cite{davis2019deep} they focused on the entity relationship detection problem in forms~\cite{FUNSD} using a simple CNN as a text line detector and then detecting key-value relationship pairs using a heuristic based on each relationship candidate score generated from the model. Sarkar et al.~\cite{sarkar2020document} rather focused on extracting the form structure by reformulating the problem as a semantic segmentation (pixel labeling) task. They used a U-Net based architectural pipeline, predicting all levels of the document hierarchy in parallel, making it quite efficient.

GNN for document understanding was first introduced for mainly key DLA sub-tasks that include table detection~\cite{riba2019table} and table structure recognition~\cite{qasim2019rethinking}. The key idea behind its introduction was to utilize the powerful geometrical characteristics of a document using GNN and then to preserve the privacy of confidential textual content (especially for administrative documents) during training, making the model language-independent and more structure-reliant as proposed in~\cite{riba2019table} for detection of tables in invoices. Carbonell et. al.~\cite{carbonell2021named} used graph convolutional networks (GCNs) to solve the entity(word) grouping, labeling and entity linking tasks for form analysis. They used the information of the bounding
boxes and word embeddings as the principal node features and do not include any visual features, while they used k-nearest neighbours (KNNs) to encode the edge information. The FUDGE~\cite{davis2021visual} framework was then developed for form understanding as an extension of~\cite{davis2019deep} to greatly improve the state-of-the-art on both the semantic entity labeling and entity linking tasks by proposing relationship pairs using the same detection CNN as in~\cite{davis2019deep}. Then a graph convolutional network (GCN) was deployed with plugged visual features from the CNN so that semantic labels for the text entities were predicted jointly with the key-value relationship pairs, as they are quite related tasks. 

Inspired by this influential prior work~\cite{davis2021visual}, we aim to propose a task-agnostic GNN-based framework called~\emph{Doc2Graph} that adapts a similar joint prediction of both the tasks, semantic entity labeling and entity linking using a node classification and edge classification module respectively. Doc2Graph is established to tackle multiple challenges ranging from KIE for form understanding to layout analysis and table detection for invoice understanding, without needing any kind of huge data pre-training and being lightweight and efficient.

%% file: tex/method.tex
In this section, we present the proposed approach. First, we describe the pre-processing step that converts document images into graphs. Then, we describe the GNN model designed to tackle different kinds of tasks.

\subsection{Documents graph structure}
\label{sec:document-graph}
A graph is a structure made of nodes and edges. A graph can be seen as a language model representing a document in terms of its segments (text units) and relationships. A preprocessing step is required. Depending on the task, different levels of granularity have to be considered for defining the constituent objects of a document. They can be single words or entities, that is, groups of words that share a certain property (e.g., the name of a company). In our work we try both as the starting point of the pipeline: we apply an OCR to recognize words, while a pre-trained object detection model for detecting entities.
The chosen objects, once found, constitute the nodes of the graph.

At this point, nodes need to be connected through edges. Finding the optimal set of edges to create the graph is anything but trivial: usually in literature heuristics are applied, e.g. using a visibility graph~\cite{riba2019table}. These approaches: (i) do not generalise well on different layouts; (ii) strongly rely on the previous node detection processes, which are often prone to errors; (iii) generate noise in the connections, since bounding box of objects could cut out important relations or allow unwanted ones; (iv) exclude in advance sets of solutions, e.g. answers far from questions.
To avoid those behaviours, we do not make any assumption a priori on the connectivity: we build a fully connected graph and we let the network learn by itself what relations are relevant.

\subsection{Node and edge features}
\label{sec:features}
In order to learn, suitable features should be associated to nodes and edges of the graph. In documents, this information can be extracted from sources of different modalities, such as visual, language and layout ones.
Different methods can be applied to encode a node (either word or entity) to enrich its representation. In our pipeline, with the aim to possibly keep it lightweight, we include:
\begin{itemize}
    \item a language model to encode the text. We use the spaCy large English model to get word vector representations of words and entities;
    \item a visual encoder to represent style and formatting. We pretrain a U-Net\cite{DBLP:journals/corr/RonnebergerFB15} on FUNSD for entities segmentation. Since U-Net uses feature maps at different encoder's layers to segment the images, we decide to use all these information as visual features. Moreover, it is important to highlight that, for each features map, we used a RoI Alignment layer to extract the features relative to each entities bounding box;
    \item the absolute normalized positions of objects inside a document; layout and structure are meaningful features to include in industrial documents, e.g. for key-value associations.
\end{itemize}
As for the edges, to the best of our knowledge, we propose two new sets of features to help both the node and the edge classification tasks:
\begin{itemize}
    \item a normalized euclidean distance between nodes, by means of the minimum distance between bounding boxes. Since we are using a fully connected graph this is crucial for the aggregation node function in use to keep locality property during the message passing algorithm;
    \item relative positioning of nodes using polar coordinates. Each source node is considered to be in the center of a Cartesian plane and all its neighbors are encoded by means of distance and angle. We discretize the space into bins (one-hot encoded), which number can be chosen, instead of using normalized angles: a continuous representation of the angle is challenging because, for instance, two points at the same distance with angles $360^{\circ}$ and $0^{\circ}$ would be encoded differently.
\end{itemize}

\subsection{Architecture}
\label{sec:architecture}
\begin{figure}[t]
    \centering
    \includegraphics[width=\textwidth]{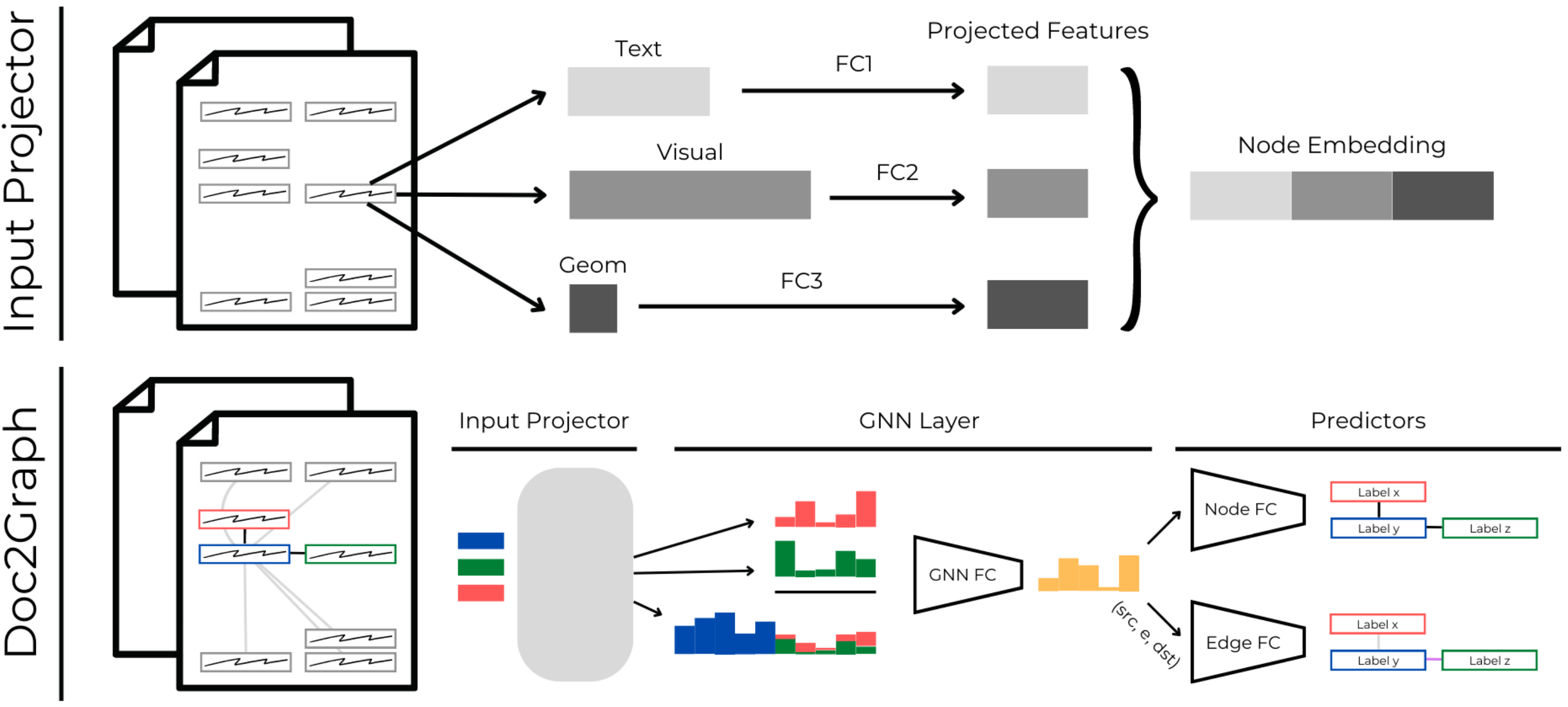}
    \caption{\textbf{Our proposed Doc2Graph framework}. For visualisation purposes, the architecture shows the perspective of one node (the blue one in Doc2Graph).}
    \label{fig:model}
\end{figure}
Each node feature vector passes through our proposed architecture (Fig. \ref{fig:model}, visualization of GNN layer inspired by \href{https://distill.pub/2021/gnn-intro/}{``A Gentle Introduction to GNNs''}): the connectivity defines the neighborhood for the message passing, while the weight learnable matrices are shared across all nodes and edges, respectively. We make use of four different components:
\begin{itemize}
    \item Input Projector: this module applies as many fully connected (FC) layers as there are different modalities in use, to project each of their representations into inner spaces of the same dimension; e.g., we found it to be not very informative combine low dimensional geometrical features with high dimensional visual ones, as they are;
    \item GNN Layer: we make use of a slightly different version of GraphSAGE \cite{hamilton2017inductive}. Using a fully connected graph, we redefine the aggregation strategy (eq. \ref{eq:graph-sage});
    \item Node Predictor: this is a FC layer, that maps the representation of each node into the number of target classes;
    \item Edge Predictor: this is a two-FC layer, that assigns a label to each edge.  To do so, we propose a novel aggregation on edges (eq. \ref{eq:edges}).
\end{itemize}

\subsubsection{GNN Layer}
\label{sec:gnn}
Our version of GraphSAGE slightly differs in the neighborhood aggregation. At layer $l$ given a node $i$, $h_i$ its inner representation and $N(i)$ its set of neighbors, the aggregation is defined as:
\begin{equation}
    h_{N(i)}^{l+1} = aggregate(\{h^{l}_{j}, \forall j \in N(i) \})
\end{equation}
where $aggregate$ can be any permutation invariant operation, e.g. sum or mean.
Usually, in other domains, the graph structure is naturally given by the data itself but, as already stated, in documents this can be challenging (sec. \ref{sec:document-graph}). Then, given a document, we redefine the above equation as:
\begin{equation}
\label{eq:graph-sage}
    h_{N(i)}^{l+1} = \frac{c}{|\Upsilon(i)|} \sum_{j \in \Upsilon(i)} h^{l}_{j}
\end{equation}
where $\Upsilon(i) = \{j \in N(i): |i - j| < threshold\}$, $|i - j|$ is the Euclidean distance of nodes $i$ and $j$ saved (normalized between 0 and 1) on their connecting edge, and $c$ is a constant scale factor.

\subsubsection{Edge Predictor}
\label{sec:edge-predictor}
We consider each edge as a triplet ($src, e, dst$): $e$ is the edge connecting the source ($src$) and destination ($dst$) node. 
The edge representation $h_e$ to feed into the two-FC classifier is defined as:
\begin{equation}
\label{eq:edges}
    h_e = h_{src}\;\Vert\;h_{dst}\;\Vert\;cls_{src}\;\Vert\;cls_{dst}\;\Vert\;e_{polar}
\end{equation}
where $h_{src}$ and $h_{dst}$ are the node embeddings output of the last GNN layer, $cls_{scr}$ and $cls_{dst}$ are the softmax of the output logits of the previous node predictor layer, $e_{polar}$ are the polar coordinates described in sec \ref{sec:features} and $\Vert$ is the concatenation operator. These choices have been made because: (i) relative positioning on edges is stronger compared to absolute positioning on nodes: the local property introduced by means of polar coordinates can be extended to different data, e.g. documents of different sizes or orientations; (ii) if the considered task comprise also the classification of nodes, their classes may help in the classification of edges, e.g. in forms it should not possible to find an answer connected to another answer. 

Given the task, graphs can be either undirected or directed: both are represented with two or one directed edge between nodes, respectively. In the first case, the order does not matter and so the above formula can be redefined as:
\begin{equation}
    h_e = (h_{src} + h_{dst})\;\Vert\;cls_{src}\;\Vert\;cls_{dst}\;\Vert\;e_{polar}
\end{equation}

%% file: tex/results.tex
In this chapter we present experiments of our method on two different datasets, FUNSD and RVL-CDIP invoices, to tackle three tasks: entity linking, layout analysis and table detection. We also discuss results compared to other methods.

\subsection{Proposed model}
We performed ablation studies on our proposed model for entity linking on FUNSD without contribution and classification of nodes (Fig. \ref{fig:model}), since we found it to be the most challenging task. In Tab. \ref{tab:ablations} we report different combinations of features and hyperparameters. Geometrical and textual features make the largest contribution, while visual features bring almost three points more to the Key-Value F1 score by an important increase in terms of network parameters (2.3 times more). Textual and geometrical features remain crucial for the task at hand, and their combination increase by a large amount both of their scores when used in isolation. This may be due to two facts: (i) our U-Net has not been included during the GNN training time (as done in \cite{davis2021visual}), unable to adjust the representation for spotting key-value relationship pairs; (ii) the segmentation task used to train the backbone do not yield useful features for that goal (as shown in Tab. \ref{tab:ablations}).
The hyperparameters shown in the table refer to the edge predictor (EP) inner layer input dimension and the input projector fully connected (IP FC) layers (per each modality) output dimension, respectively. A larger EP is much more informative for the classification of links into `none' (cut edges, meaning no relationship) or `key-value', while more dimensions for the projected modalities helped the model to better learn the importance of their contributions. These changes bring an improvement of 13 points on the key-value F1 scores, between the third and fourth line of the table where we keep the features fixed. We do not report the score relative to others network settings since their changes only brought a decrease overall metrics. We use a learning rate of $10^{-3}$ and a weight decay of $10^{-4}$, with a dropout of 0.2 over the last FC layer. The threshold over neighbor nodes and their contribution scale factor (sec. \ref{sec:gnn}) are fixed to 0.9 and 0.1, respectively. The bins to discretize the space for angles (sec. \ref{sec:edge-predictor}) are 8. We apply one GNN layer before the node and edge predictors.

\begin{table}[t]
    \centering
    \resizebox{1\textwidth}{!}{%
        \begin{tabular}{@{}lll ll lllll l@{}}
            \toprule
            \multicolumn{3}{c}{\textit{Features}}                &                     &                        & \multicolumn{2}{c}{\textit{F\textsubscript{1} per classes} ($\uparrow$)} &                                & \\ \cmidrule{1-3} \cmidrule{6-7}
            \textbf{Geometric} & \textbf{Text} & \textbf{Visual} & \textbf{EP Inner dim} & \textbf{IP FC dim} & \textbf{None}                      & \textbf{Key-Value}                 & \textbf{AUC-PR} ($\uparrow$)   & \textbf{\# Params $\times10^6$} ($\downarrow$) \\ \midrule
            \cmark        & \xmark             & \xmark          & 20                  & 100                    & 0.9587                             & 0.1507                        & 0.6301                         & 0.025 \\
            \xmark        & \cmark             & \xmark          & 20                  & 100                    & 0.9893                             & 0.1981                        & 0.5605                         & 0.054 \\
            \cmark        & \cmark             & \xmark          & 20                  & 100                    & 0.9941                             & 0.4305                        & 0.7002                         & 0.120 \\
            \cmark        & \cmark             & \xmark          & 300                 & 300                    & 0.9961                             & 0.5606                        & 0.7733                         & 1.18 \\
            \cmark        & \cmark             & \cmark          & 300                 & 300                    & \textbf{0.9964}                    & \textbf{0.5895}               & \textbf{0.7903}                & 2.68 \\
            \bottomrule \\ 
        \end{tabular}%
    }
    \caption{\textbf{Ablation studies of Doc2Graph model}. EP Inner dim and IP FC dim show edge predictor layer input dimension and the input projector fully connected layers output dimension, respectively. AUC-PR refers to the key-value edge class. The \# Params refers to Doc2Graph trainable parameters solely.}
    \label{tab:ablations}
\end{table}

\subsection{FUNSD}
\subsubsection{Dataset}
The dataset\cite{FUNSD} comprises 199 real, fully annotated, scanned forms. The documents are selected as a subset of the larger RVL-CDIP\cite{RVL-CDIP} dataset, a collection of 400,000 grayscale images of various documents. The authors define the Form Understanding (FoUn) challenge into three different tasks: word grouping, semantic entity labeling and entity linking.
A recent work \cite{revised-FUNSD} found some inconsistency in the original labeling, which impeded its applicability to the key-value extraction problem. In this work, we are using the revised version of FUNSD.
\begin{figure}[!b]
    \centering
    \includegraphics[width=.65\textwidth]{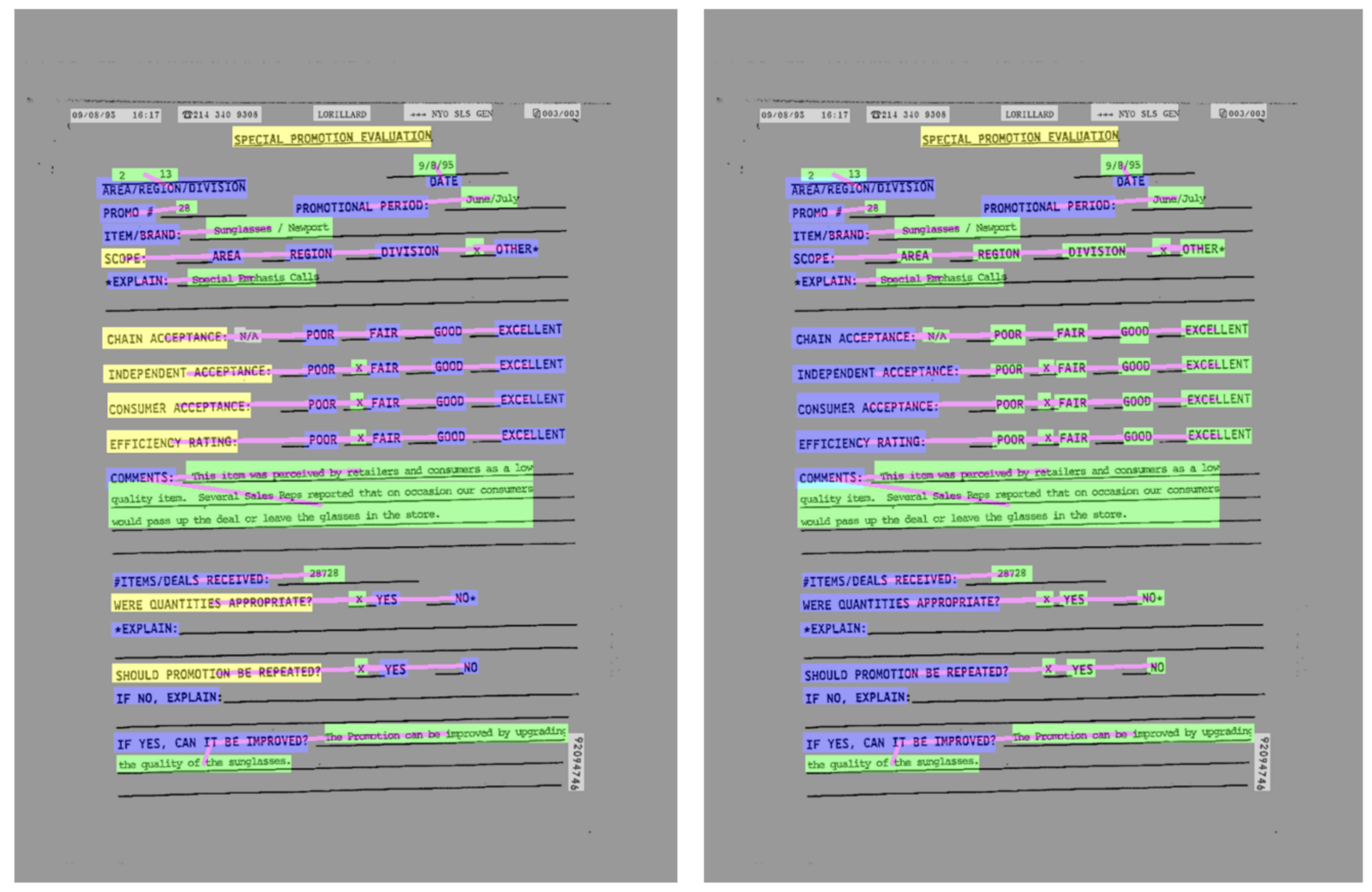}
    \caption{Image taken from \cite{revised-FUNSD}: the document on the right is the revised version of the document on the left, where some answers (green) are mislabeled as question (blue), and some questions (blue) are mislabeled as headers (yellow)}
    \label{fig:revised-funsd}
\end{figure}

\subsubsection{Entity Detection}
\begin{figure}[t]
    \centering
    \includegraphics[width=.8\textwidth]{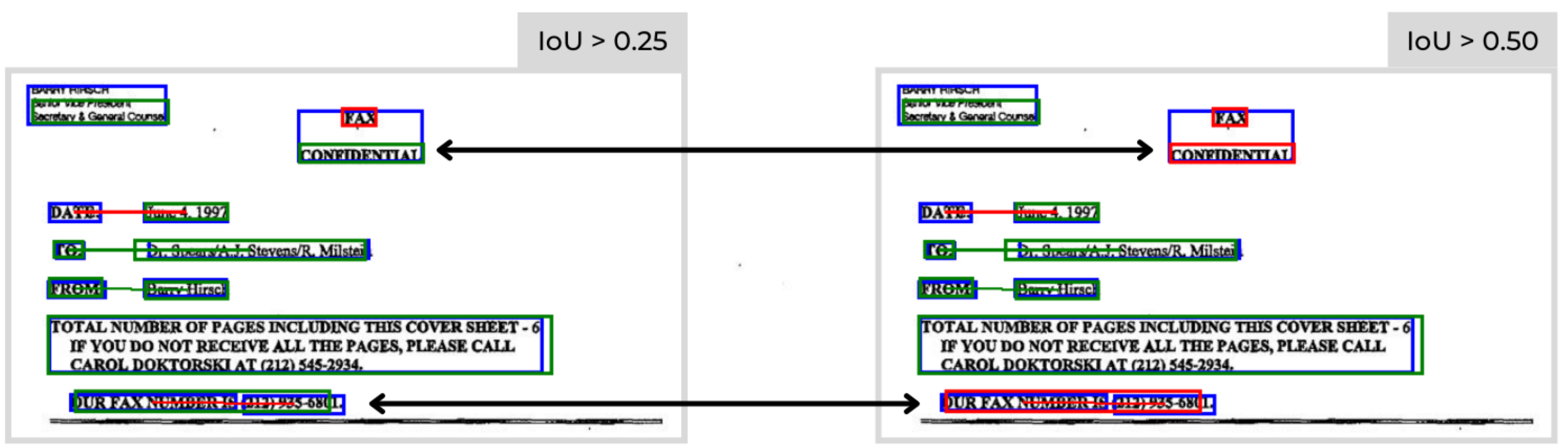}
    \caption{Blue boxes are FUNSD entities ground truth, green boxes are the correct detected one (with IoU $> 0.25 / 0.50$), while red boxes are the false positive ones.}
    \label{fig:det}
\end{figure}

Our focus is on the GNN performances but, for comparison reasons, we used a YOLOv5 small \cite{YOLOv5} to detect entities (pretrained on COCO \cite{lin2014microsoft}). In \cite{FUNSD} the word grouping task is evaluated using the ARI metric: since we are not using words, we evaluated the entity detection with F1 score using two different IoU thresholds (Tab. \ref{tab:iou_f1}). For the semantic entity labeling and entity linking tasks we use IoU $>0.50$ as done in \cite{davis2021visual}: we did not perform any optimization on the detector model, which introduces a high drop rate for both entities and links. We create the graphs on top of YOLO detections, linking the ground truth accordingly (Fig. \ref{fig:det}): false positive entities (red boxes) are labeled as class 'other', while false negative entities cause some key-value pairs to be lost (red links). The new connections created as a consequence of wrong detections are considered false positives and labeled as `none'. 

\begin{table}[t!]
    \centering
    \resizebox{0.7\textwidth}{!}{%
        \begin{tabular}{@{}lllllll@{}}
            \toprule
                            & \multicolumn{3}{c}{\textit{Metrics} ($\uparrow$)}                 && \multicolumn{2}{c}{\textit{\% Drop Rate} ($\downarrow$)} \\ \cmidrule{2-4} \cmidrule{6-7}
            \textbf{IoU}    & \textbf{Precision} & \textbf{Recall} & \textbf{F\textsubscript{1}} && \textbf{Entity}            & \textbf{Link}               \\ \midrule
            0.25            & 0.8728            & 0.8712          & 0.8720                      && 12.72                      & 16.63                       \\
            0.50            & 0.8132            & 0.8109          & 0.8121                      && 18.67                      & 25.93                       \\
            \bottomrule \\
        \end{tabular}%
    }
    \caption{\textbf{Entity detection results}. YOLOv5 \cite{YOLOv5}-small performances on the entity detection task.}
    \label{tab:iou_f1}
\end{table}

\begin{table}[b!]
    \centering
    \resizebox{\textwidth}{!}{%
        \begin{tabular}{@{}lllll@{}}
            \toprule
                                                       &              & \multicolumn{2}{c}{\textit{F\textsubscript{1}} ($\uparrow$)}                            &                                                \\ \cmidrule{3-4} 
            \textbf{Method}                            & \textbf{GNN} & \textbf{Semantic Entity Labeling}                     & \textbf{Entity Linking}  & \textbf{\# Params $\times10^6$ ($\downarrow$)} \\ \midrule
            BROS \cite{hong2020bros}                   & \xmark       & \textbf{0.8121}                                                       & \textbf{0.6696}           & 138                                           \\
            LayoutLM \cite{xu2020layoutlm,hong2020bros}& \xmark       & 0.7895                                                       & 0.4281                    & 343                                           \\ \\
            
            FUNSD \cite{FUNSD}                         & \cmark       & 0.5700                                                       & 0.0400                   & -                                              \\
            Carbonell et al. \cite{carbonell2021named} & \cmark       & 0.6400                                                       & 0.3900                   & 201                                            \\
            FUDGE w/o GCN \cite{davis2021visual}       & \xmark       & 0.6507                                                       & 0.5241                   & 12                                             \\
            FUDGE \cite{davis2021visual}               & \cmark       & 0.6652                                                       & 0.5662                   & 17                                             \\ \\
            
            Doc2Graph + YOLO                           & \cmark       & 0.6581 $\pm$ 0.006                                           & 0.3882 $\pm$ 0.028       & 13.5                                           \\
            Doc2Graph + GT                             & \cmark       & \underline{0.8225} $\pm$ 0.005                                  & 0.5336 $\pm$ 0.036       & \textbf{6.2 }                                  \\
            \bottomrule \\
        \end{tabular}%
    }
    \caption{\textbf{Results on FUNSD}. The results have been shown for both semantic entity labeling and entity linking tasks with their corresponding metrics.}
    \label{tab:funsd_comparison}
\end{table}

\subsubsection{Numerical results}
We trained our architecture (sec. \ref{sec:architecture}) with a 10-fold cross validation. Since we found high variance in the results, we report both mean and variance over the 10 best models chosen over their respective validation sets. The objective function in use ($L$) is based on both node ($L_n$) and edge ($L_e$) classification tasks: $L = L_n + L_e$.
In Tab. \ref{tab:funsd_comparison} we report the performances of our model Doc2Graph compared to other language models \cite{hong2020bros,xu2020layoutlm} and graph-based techniques \cite{carbonell2021named,davis2021visual}. The number of parameters \# Params refer to the trainable Doc2Graph pipeline (that includes the U-Net and YOLO backbones); for the spaCy word-embedding details, refer to \href{https://spacy.io/models/en\#en_core_web_lg}{their documentation}. Using YOLO our network outperforms \cite{carbonell2021named} for semantic entity labeling and meets their model on entity linking, using just 13.5 parameters. We could not do better than FUDGE, which still outperforms our scores. Their backbone is trained for both tasks along with the GCN (GCN that adds just minor improvements). The gap, especially on entity linking, is mainly due to the low contributions given by our visual features (Tab. \ref{tab:ablations}) and the detector in use (Tab. \ref{fig:det}).
We also report the results of our model initialized with ground truth (GT) entities, to show how it would perform in the best case scenario. Entity linking remains a harder task compared to semantic entity labeling and only complex language models seem to be able to solve it. Moreover, for the sake of completeness, we highlight that, with good entity representations, our model outperforms all the considered architectures for the Semantic Entity Labeling task. Finally, we want to further stress that the main contribution of a graph-based method is to yield a simpler but more lightweight solution.

\subsubsection{Qualitative results}
\begin{figure}[t!]
    \centering
    \begin{subfigure}[b]{0.45\textwidth}
        \centering
        \includegraphics[width=0.9\textwidth]{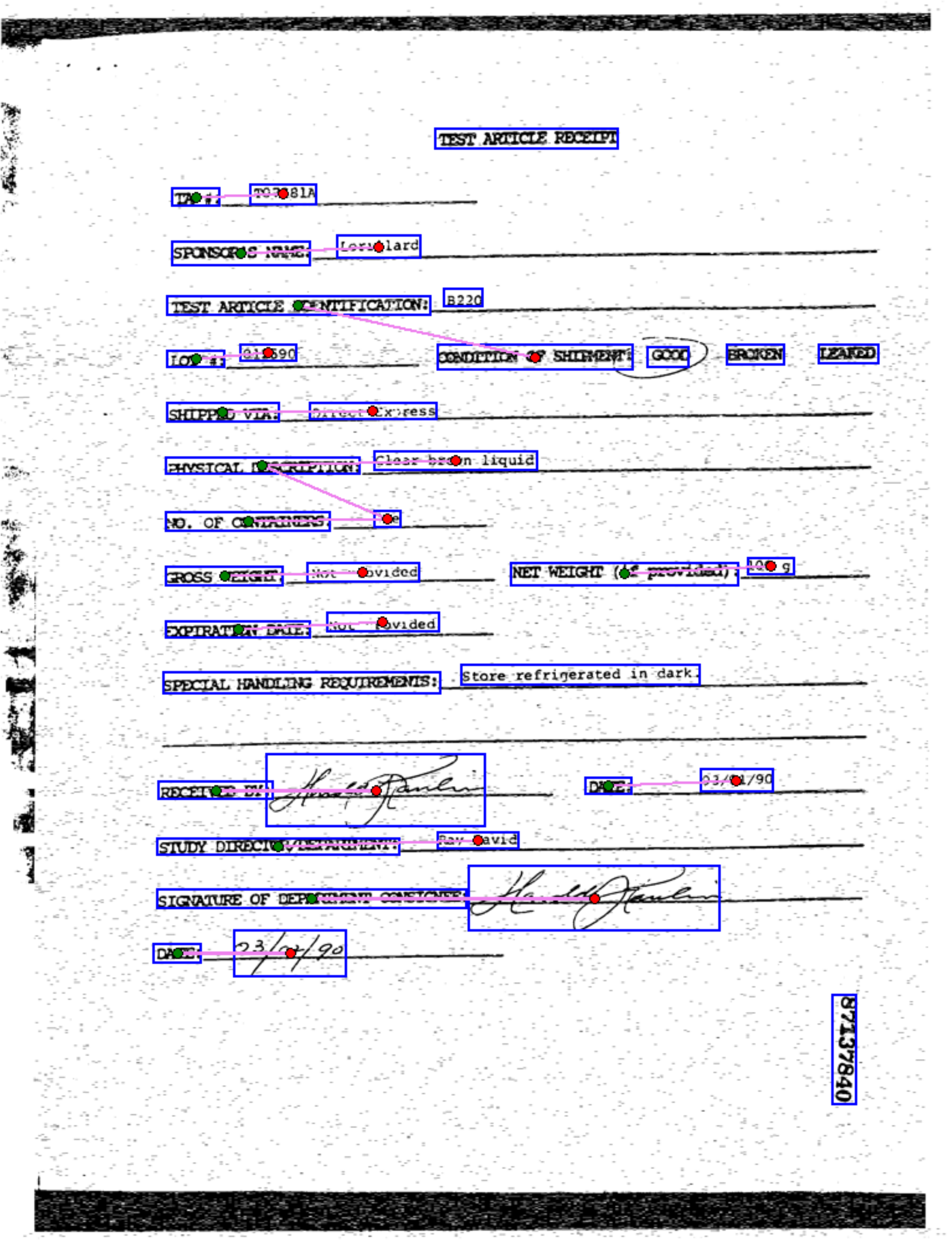}
        \caption{}
        \label{fig:case_a}
    \end{subfigure}
    \begin{subfigure}[b]{0.45\textwidth}
        \centering
        \includegraphics[width=0.9\textwidth]{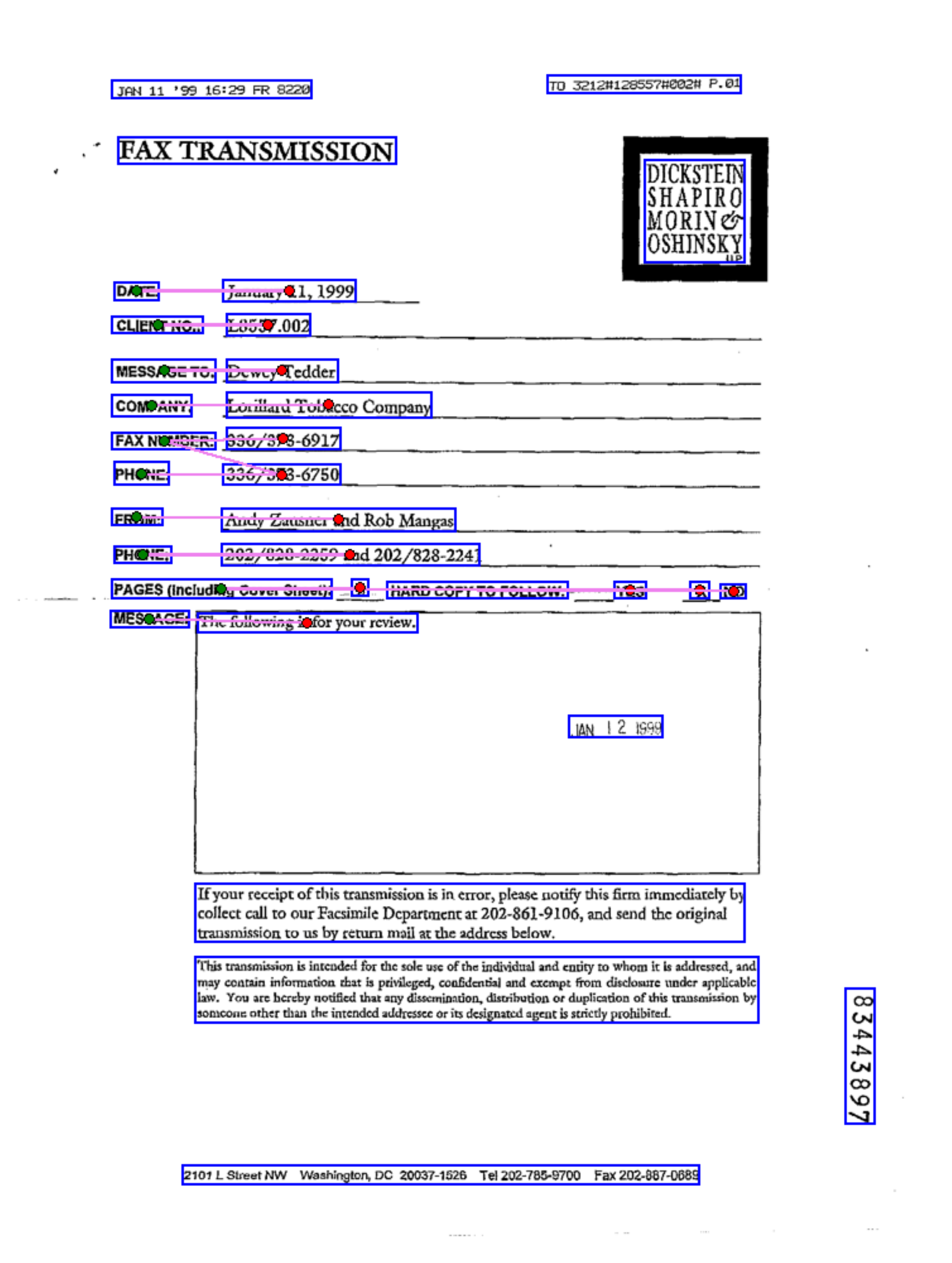}
        \caption{}
        \label{fig:case_b}
    \end{subfigure}
    \begin{subfigure}[b]{0.45\textwidth}
        \centering
        \includegraphics[width=0.9\textwidth]{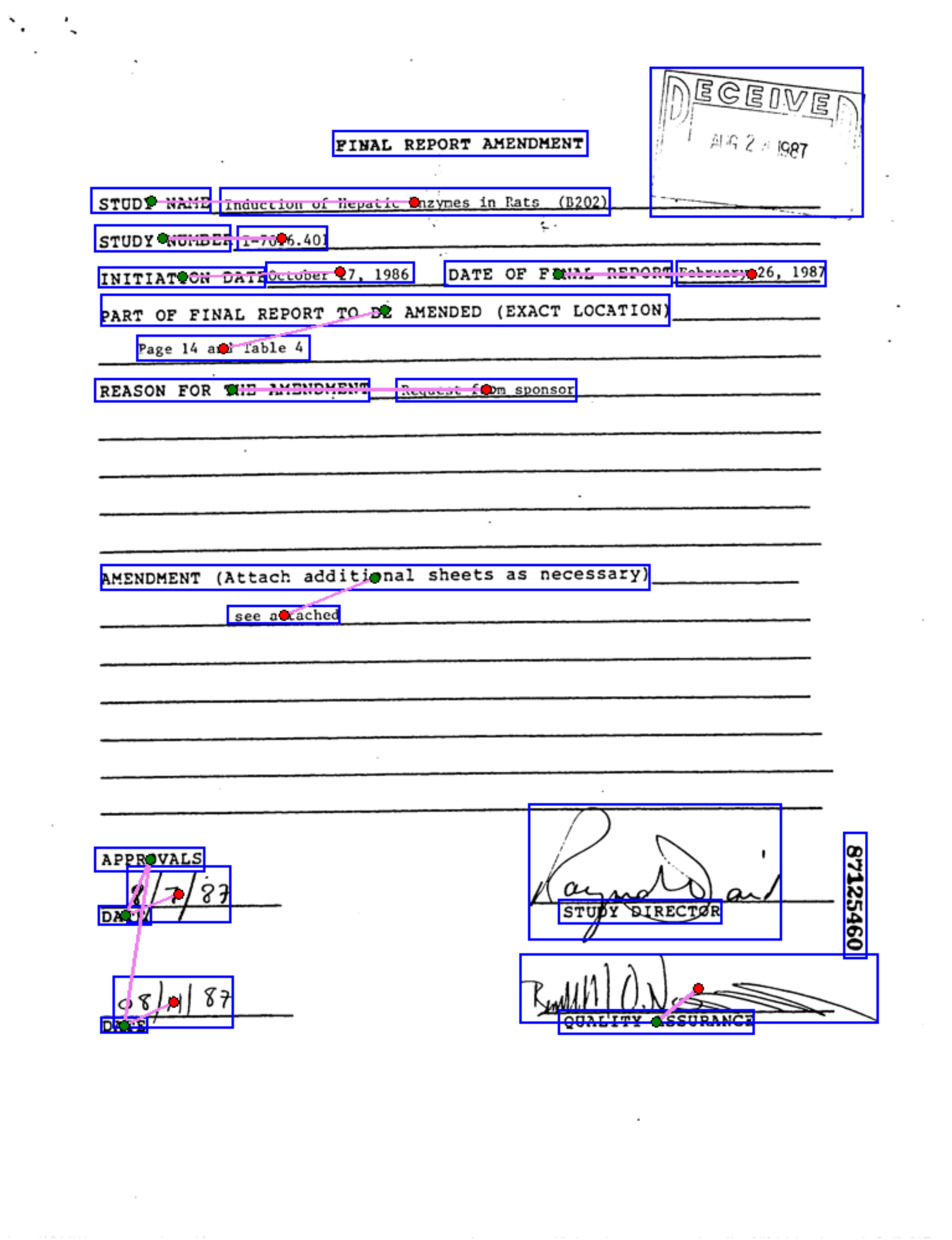}
        \caption{}
        \label{fig:case_c}
    \end{subfigure}
    \begin{subfigure}[b]{0.45\textwidth}
        \centering
        \includegraphics[width=0.9\textwidth]{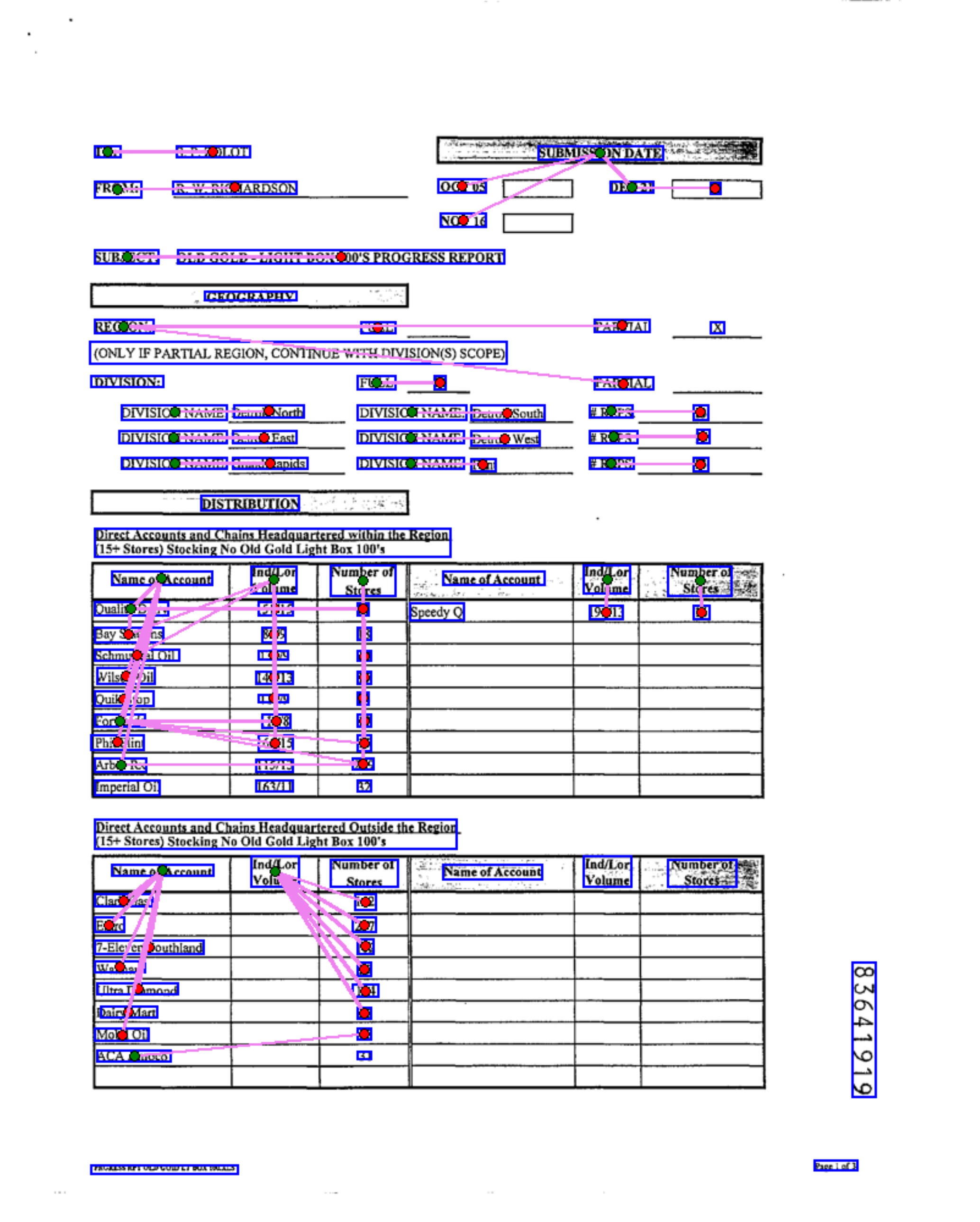}
        \caption{}
        \label{fig:case_d}
    \end{subfigure}
    \caption{\textbf{Entity Linking on FUNSD}. We make use of directed edges: green and red dots mean source and destination nodes, respectively.}
    \label{fig:link-prediction}
\end{figure}

The order matters for detecting key-value relationship, since the direction of a link induce a property for the destination entity that enrich its meaning. Differently from FUDGE~\cite{davis2021visual} we do make use of directed edges, which led to a better understanding of the document having interpretable results. In Fig. \ref{fig:link-prediction} we show our qualitative results using Doc2Graph on groundtruth: green and red dots mean source and destination nodes, respectively. As shown in the different example cases, Fig.~\ref{fig:link-prediction}\subref{fig:case_a} and ~\ref{fig:link-prediction}\subref{fig:case_b} resemble a simple structured form layout with directed one-to-one key-value association pairs and Doc2Graph manages to extract them. On the contrary, where the layout appears to be more complex as in Fig.~\ref{fig:link-prediction}\subref{fig:case_d}, Doc2Graph fails to generalize the concept of one-to-many key-value relationship pairs. This may be due to the small number of trainable samples we had in our training data and the fact that header-cells usually present different positioning and semantic meaning. In the future we will integrate a table structure recognition path into our pipeline, hoping to improve the extraction of all kinds of key-value relationships in such more complex layout scenarios.

\subsection{RVL-CDIP Invoices}
\subsubsection{Dataset} 
\begin{figure}
     \centering
     \begin{subfigure}[b]{0.4\textwidth}
         \centering
         \includegraphics[width=0.8\textwidth]{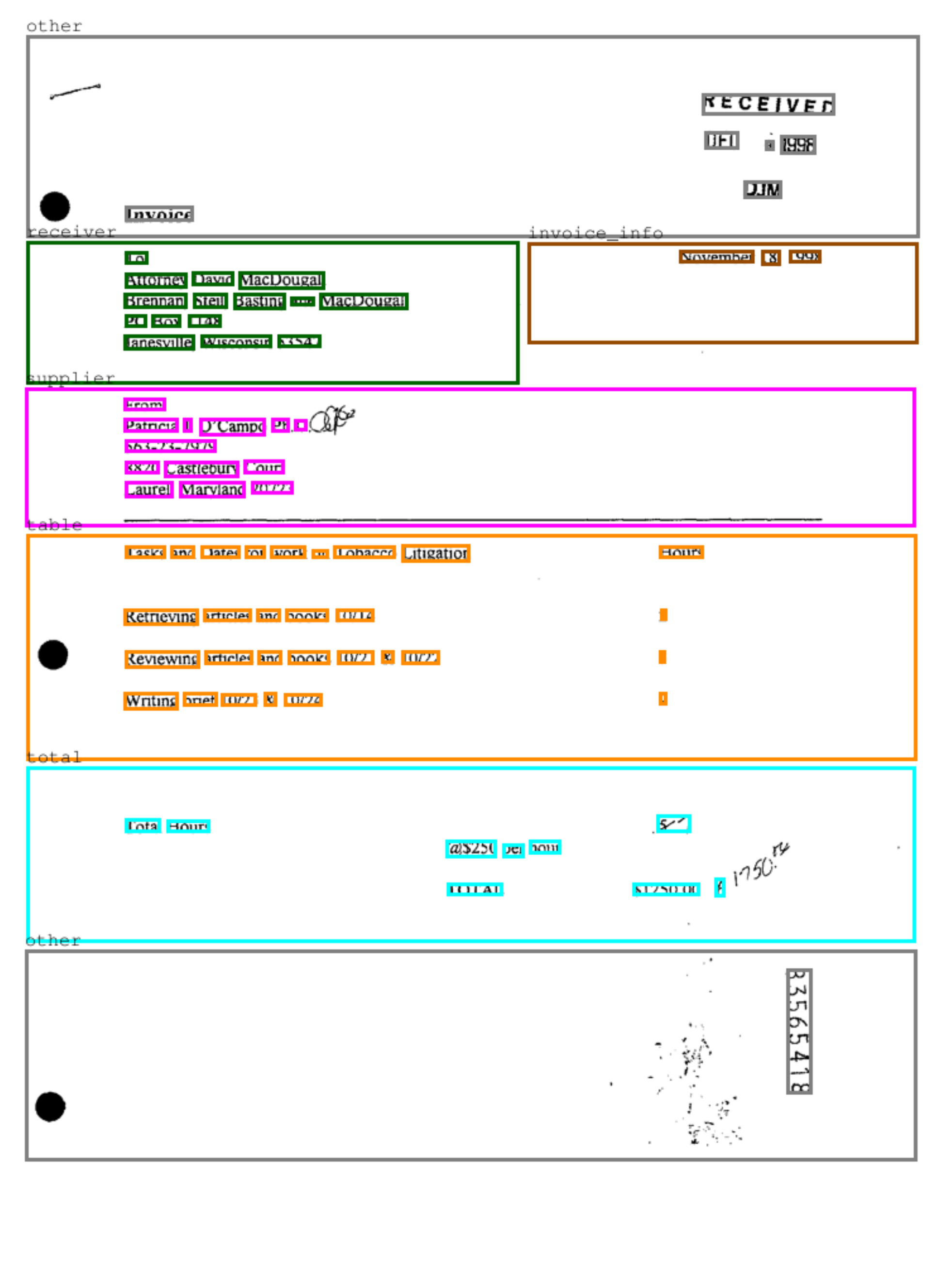}
         \label{fig:pa}
     \end{subfigure}
     \begin{subfigure}[b]{0.4\textwidth}
         \centering
         \includegraphics[width=0.8\textwidth]{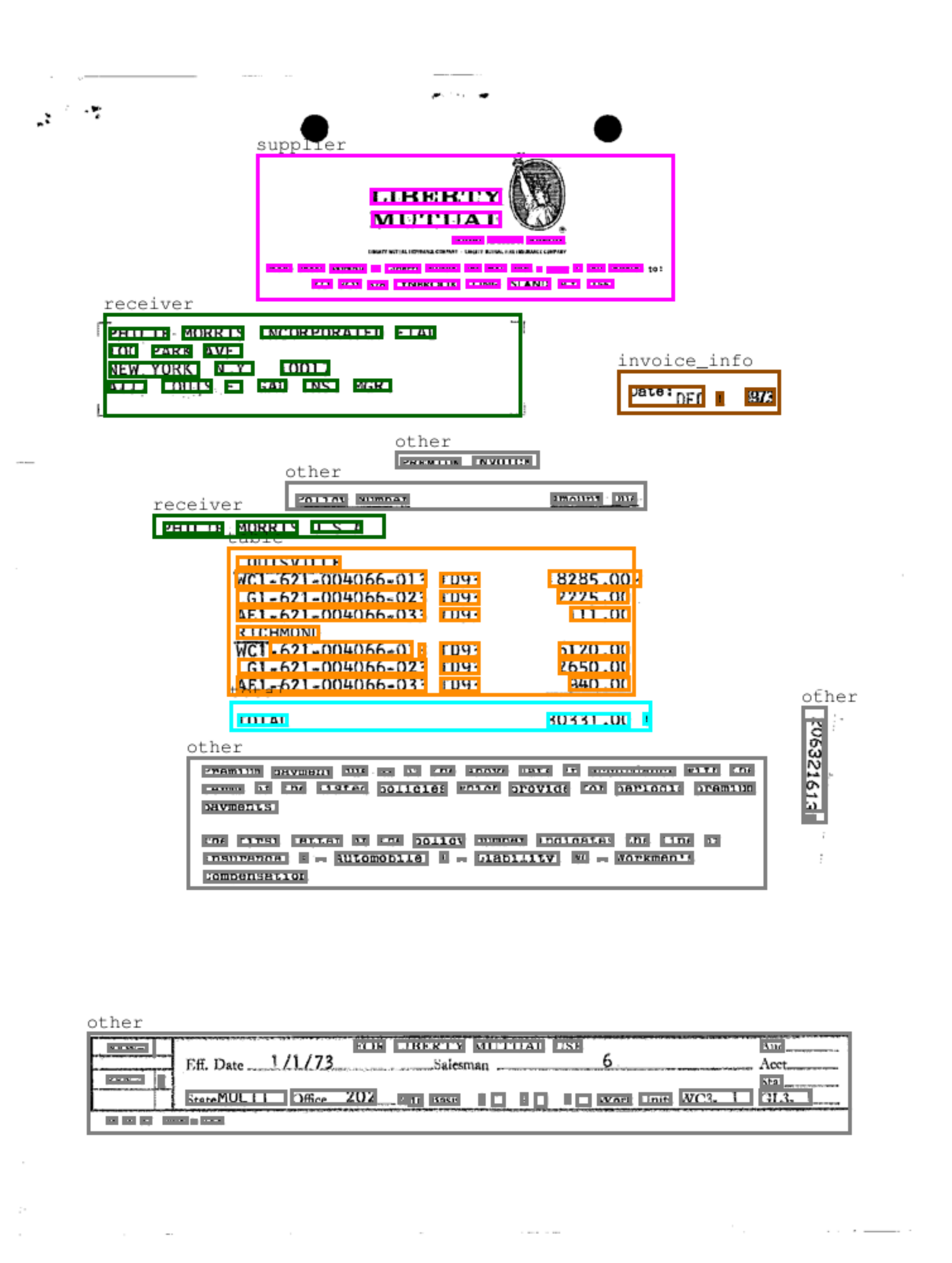}
         \label{fig:three sin x}
     \end{subfigure}
     \caption{\textbf{RVL-CDIP Invoices benchmark in \cite{riba2019table}}. There are 6 regions: supplier (pink), invoice\_info (brown), receiver (green), table (orange), total (light blue), other (gray).}
     \label{fig:invoices}
\end{figure}

In the work of Riba et al. \cite{riba2019table} another subset of RVL-CDIP has been released. The authors selected 518 documents from the invoices classes, annotating 6 different regions (two examples of annotations are shown in Fig. \ref{fig:invoices}). The task that can be performed are layout analysis, in terms of node classification, and table detection, in terms of bounding box (IoU $>$ 50). 

\subsubsection{Numerical results}
As done previously, we perform a k-fold cross validation keeping, for each fold, the same amount of test (104), val (52) and training documents (362). This time we applied an OCR to build the graph. There are two tasks: layout analysis, in terms of accuracy, and table detection, using F1 score and $IoU > 0.50$ for table regions. Our model outperforms \cite{riba2019table} in both tasks, as shown in tables \ref{tab:node_classification_accuracy} and \ref{tab:edge_classification_f1}. In particular, for table detection, we extracted the subgraph induced by the edge classified as `table' (two nodes are linked if they are in the same table) to extract the target region. Riba et al. \cite{riba2019table} formulated the problem as a binary classification: we report, for brevity, in Tab. \ref{tab:edge_classification_f1} the threshold on confidence score they use to cut out edges, that in our multi-class setting (`none' or `table') is implicitly set to 0.50 by the softmax.

\begin{table}[b]
    \centering
    \resizebox{0.6\textwidth}{!}{%
        \begin{tabular}{@{}l ll@{}}
            \toprule
                                              & \multicolumn{2}{c}{\textit{Accuracy} ($\uparrow$)} \\ \cmidrule{2-3}
            \textbf{Method}                   & \textbf{Max}   & \textbf{Mean}             \\ \midrule
            Riba et al. \cite{riba2019table}  & 62.30          & -                         \\
            Doc2Graph + OCR                   & \textbf{69.80} & \textbf{67.80} $\pm$ 1.1  \\
            \bottomrule \\
        \end{tabular}%
    }
    \caption{\textbf{Layout analysis results on RVL-CDIP Invoices}. Layout analysis accuracy scores depicted in terms of node classification task.}
    \label{tab:node_classification_accuracy}
\end{table}

\begin{table}[ht!]
    \centering
    \resizebox{\textwidth}{!}{%
        \begin{tabular}{@{}l l lll@{}}
            \toprule
                                                 &                    & \multicolumn{3}{c}{\textit{Metrics} ($\uparrow$)}                            \\ \cmidrule{3-5}
            \textbf{Method}                      & \textbf{Threshold} & \textbf{Precision}         & \textbf{Recall}   & \textbf{F\textsubscript{1}} \\ \midrule
            Riba et al. \cite{riba2019table}     & 0.1                & 0.2520                     & \textbf{0.3960}   & 0.3080                      \\
            Riba et al. \cite{riba2019table}     & 0.5                & 0.1520                     & 0.3650            & 0.2150                      \\
            Doc2Graph + OCR                      & 0.5                & \textbf{0.3786} $\pm$ 0.07 & 0.3723 $\pm$ 0.07 & \textbf{0.3754} $\pm$ 0.07  \\
            \bottomrule \\
        \end{tabular}%
    }
    \caption{\textbf{Table Detection in terms of F1 score}. A table is considered correctly detected if its IoU is greater than 0.50. Threshold values refers to the scores an edges has to have in order to do not be cut: in our case is set to 0.50 by the softmax in use.}
    \label{tab:edge_classification_f1}
\end{table}

\subsubsection{Qualitative results}
\begin{figure}[t!]
     \centering
     \begin{subfigure}[b]{0.45\textwidth}
         \centering
         \includegraphics[width=0.9\textwidth]{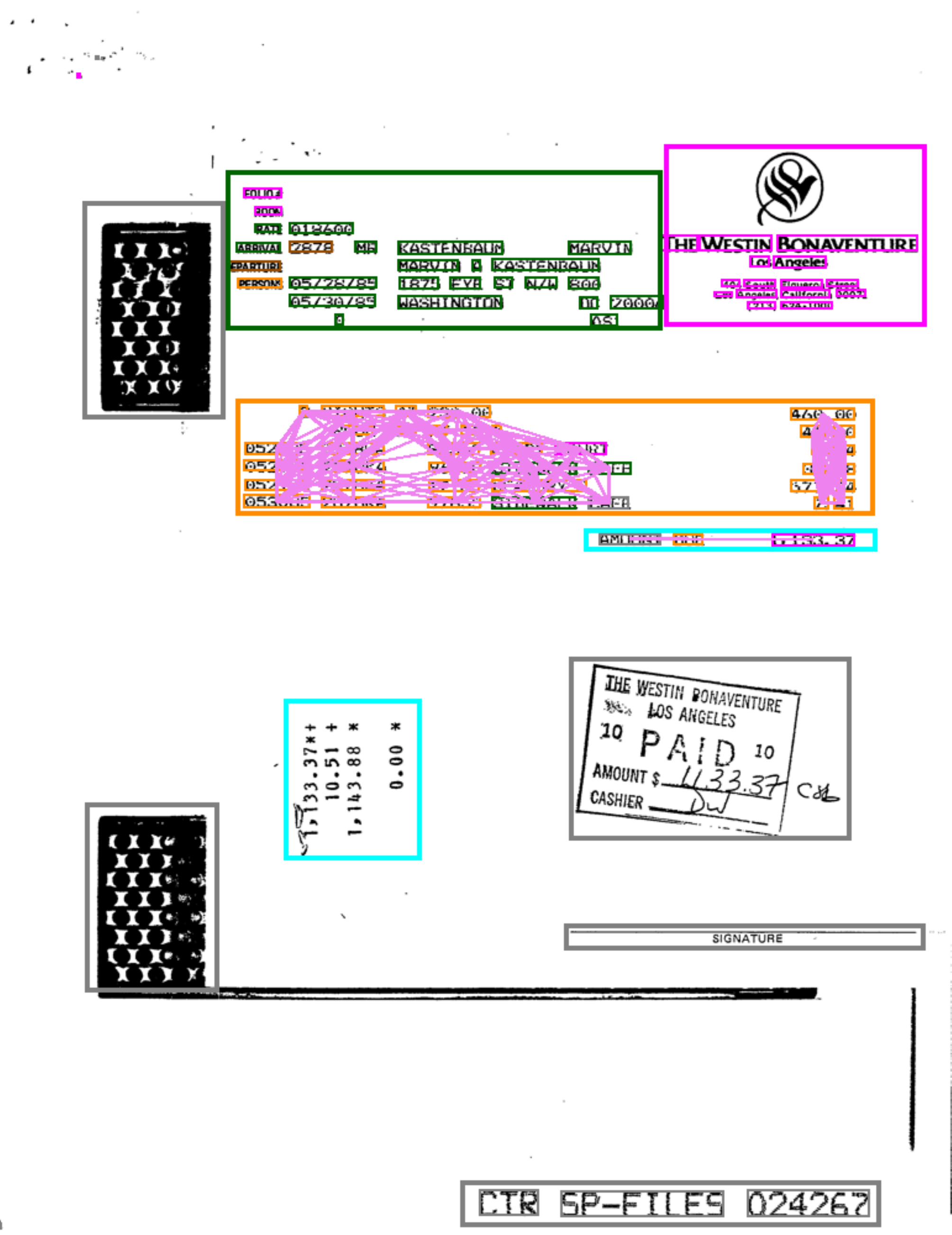}
         \caption{}
         \label{fig:rvl-a}
    \end{subfigure}
     \begin{subfigure}[b]{0.45\textwidth}
         \centering
         \includegraphics[width=0.9\textwidth]{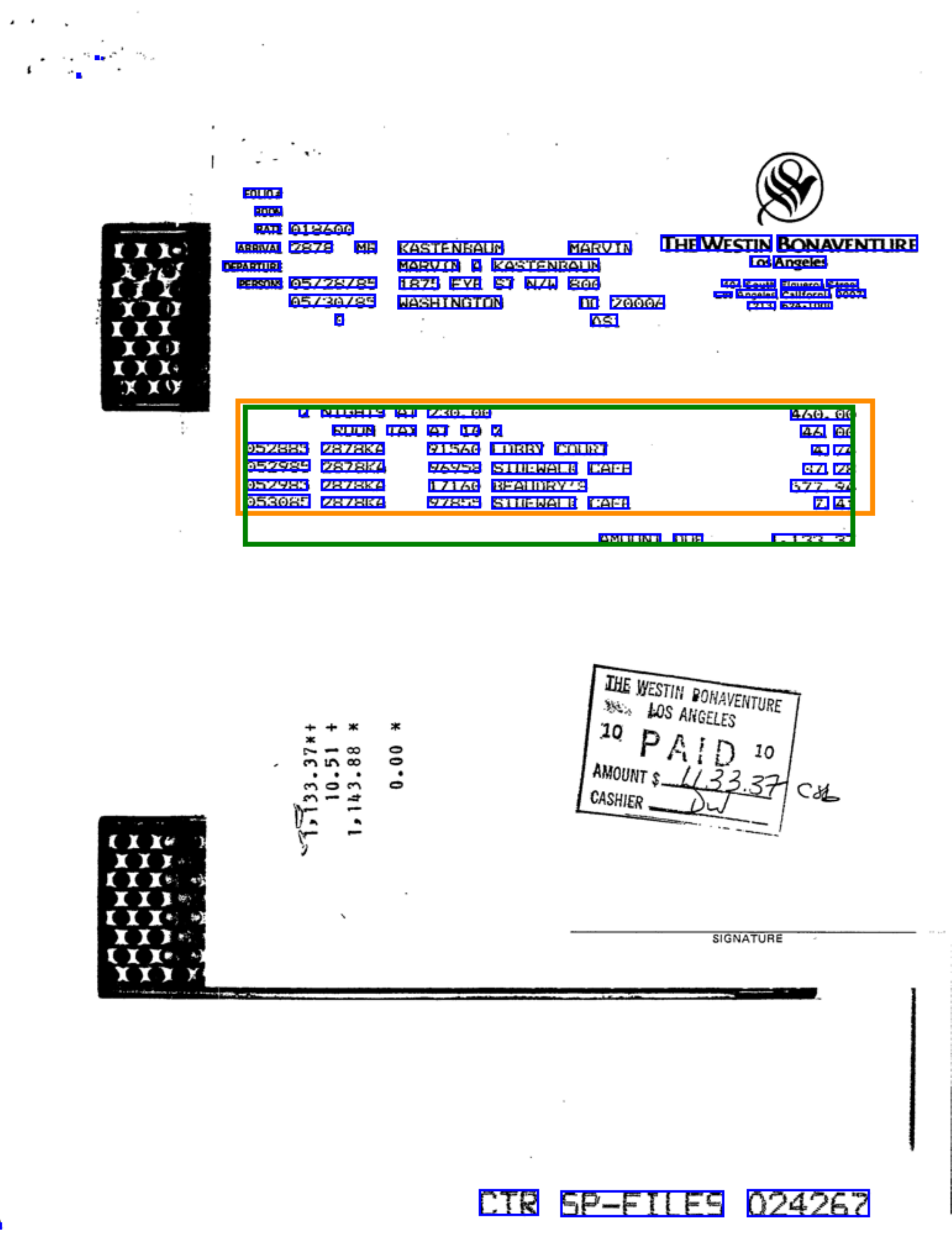}
         \caption{}
         \label{fig:rvl-b}
    \end{subfigure}
     \hfill
     \begin{subfigure}[b]{0.45\textwidth}
         \centering
         \includegraphics[width=0.9\textwidth]{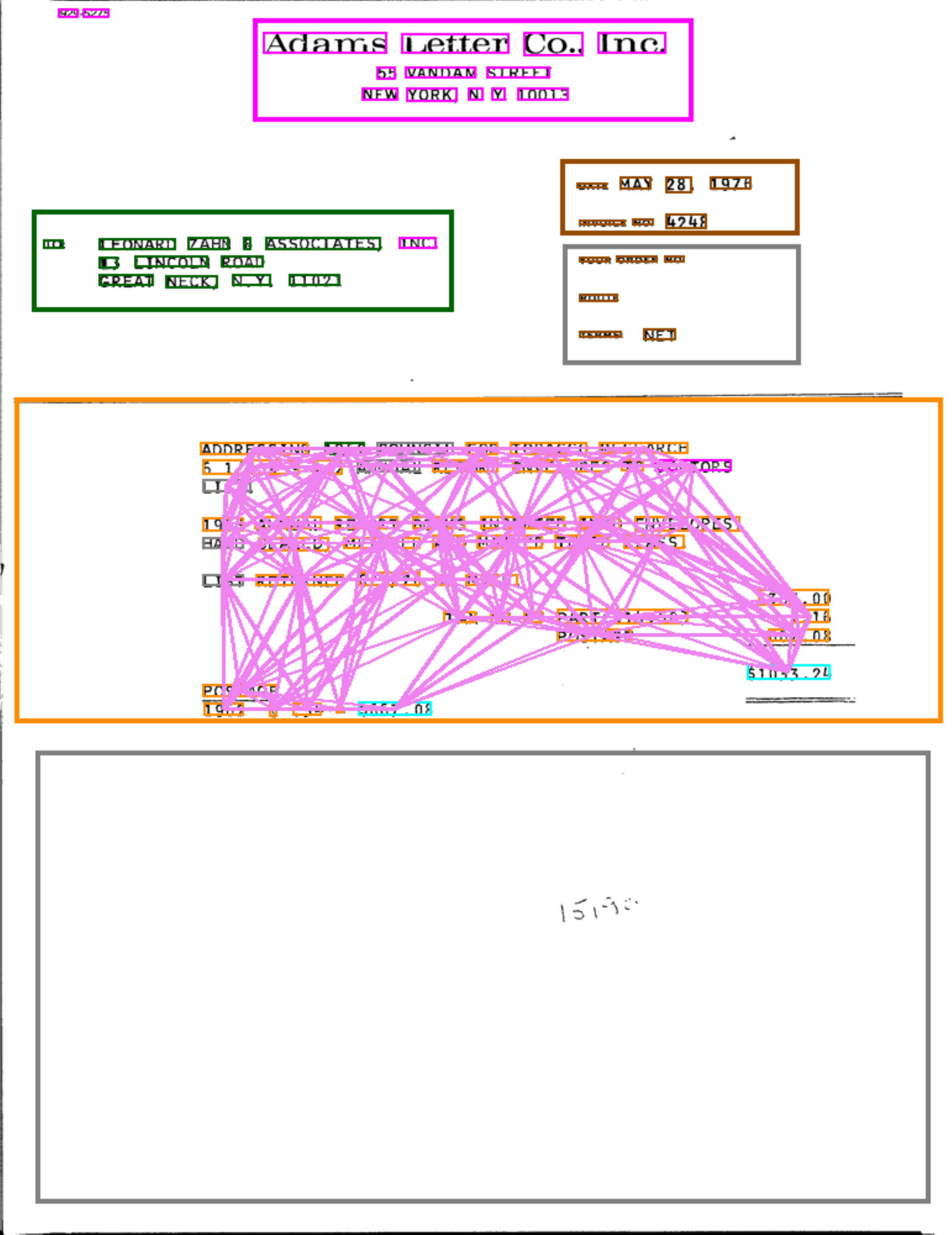}
         \caption{}
         \label{fig:rvl-c}
     \end{subfigure}
     \begin{subfigure}[b]{0.45\textwidth}
         \centering
         \includegraphics[width=0.9\textwidth]{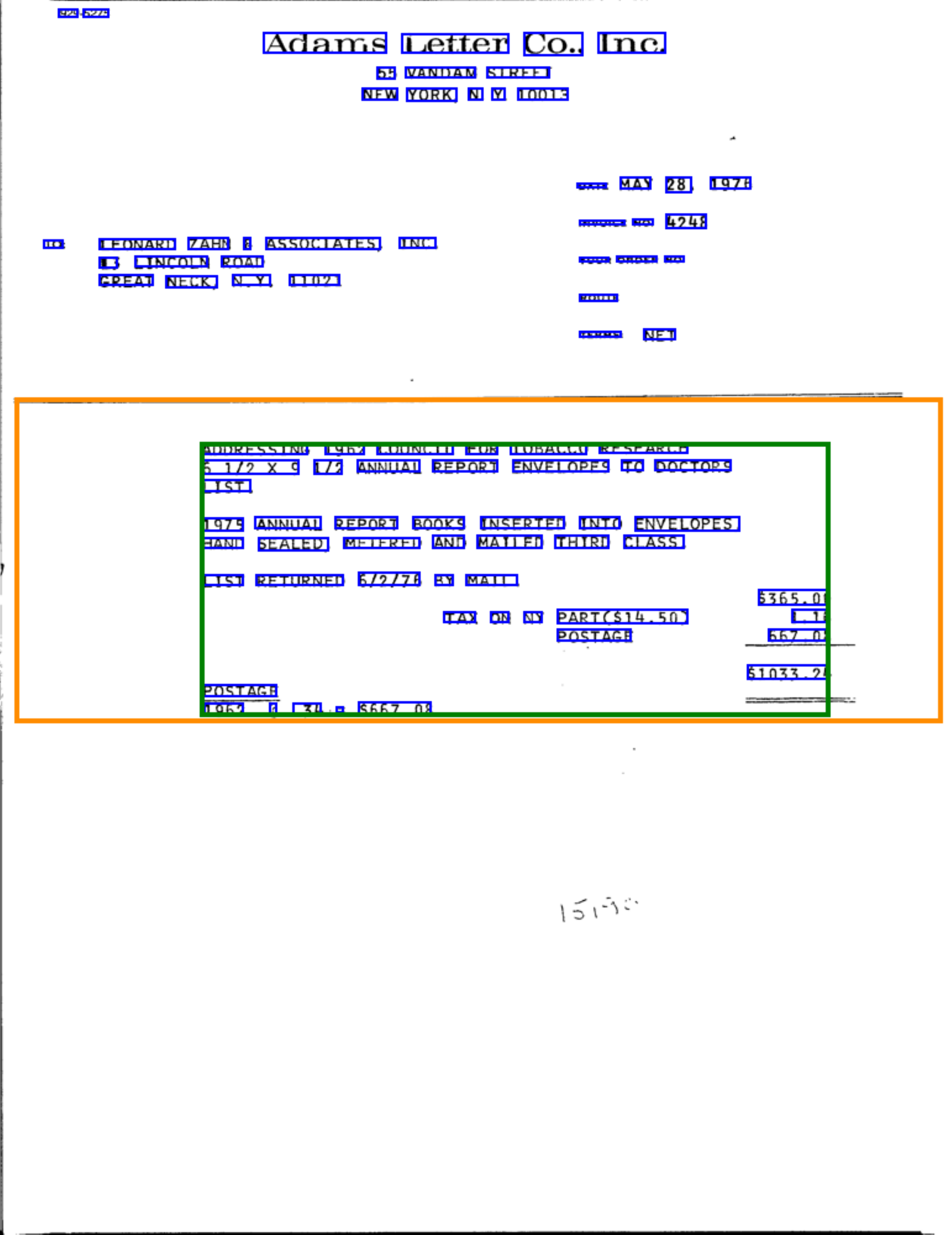}
         \caption{}
         \label{fig:rvl-d}
     \end{subfigure}
        \caption{\textbf{Layout Analysis on RVLCDIP Invoices}. Inference over two documents from RVL-CDIP Invoices, showing both Layout Analysis (a,c) and Table Detection (b,d) tasks.}
        \label{fig:invoices-pred}
\end{figure}

In Fig. \ref{fig:invoices-pred} we show the qualitative results. The two documents are duplicated to better visualize the two tasks. For layout analysis, the greater boxes colors indicate the true label that the word inside should have (the colors reflects classes as shown in Fig. \ref{fig:invoices}). For the table detection we use a simple heuristic: we take the enclosing rectangle (green) of the nodes connected by `table' edges, then we evaluate the IoU with target regions (orange). This heuristic is effective but simple and so error-prone: if a false positive is found outside table regions this could lead to a poor detection result, e.g. a bounding box including also 'sender item'  entity or 'receiver item' entity. In addition, as inferred from Figs.~\ref{fig:invoices-pred}\subref{fig:rvl-a} and ~\ref{fig:invoices-pred}\subref{fig:rvl-b}, 'total' regions could be taken out.
In the future, we will refine this behaviour by both boosting the node classification task and including 'total' as a table region for the training of edges.

%% file: tex/conclusion.tex
In this work, we have presented a task-agnostic document understanding framework based on a Graph Neural Network. We propose a general representation of documents as graphs, exploiting fully connectivity between document objects and letting the network automatically learn meaningful pairwise relationships. Node and edge aggregation functions are defined by taking into account the relative positioning of document objects. We evaluated our model on two challenging benchmarks for three different tasks: entity linking on forms, layout analysis on invoices and table detection. Our preliminary results show that our model can achieve promising results, keeping the network dimensionality considerably low.
For future works, we will extend our framework to other documents and tasks, to deeper investigate the generalization property of the GNN. We would like to explore more extensively the contribution of different source features and how to combine them in more meaningful and learnable ways.